\documentclass[runningheads]{llncs}
\usepackage[T1]{fontenc}
\usepackage{graphicx}
\usepackage{amsmath,amssymb}
\usepackage{subfigure}
\usepackage[ruled]{algorithm2e}
\usepackage{setspace}

\usepackage{threeparttable}
\usepackage{booktabs}
\usepackage{multirow}
\DeclareMathOperator*{\argmax}{arg\,max}
\newtheorem{theo}{Assumption}
\newtheorem{defi}{Definition}

\begin{document}
\title{MaxGNR: A Dynamic Weight Strategy via Maximizing Gradient-to-Noise Ratio for Multi-Task Learning}
\titlerunning{A Dynamic Weight Strategy via Maximizing Gradient-to-Noise Ratio}
%
\author{Caoyun Fan \inst{1,\dagger} \and Wenqing Chen \inst{2,\dagger} \and Jidong Tian \inst{1} \and Yitian Li \inst{1} \and Hao He \inst{1,*} \and Yaohui Jin \inst{1}}
\authorrunning{Fan et al.}
%


\institute{Shanghai Jiao Tong University, Shanghai, China  \\
\email{\{fcy3649, frank92, yitian\_li, hehao, jinyh\}@sjtu.edu.cn} \\ \and
Sun Yat-sen University, Guangzhou, China  \\
\email{chenwq95@mail.sysu.edu.cn}}

%
\maketitle              

\begin{abstract}

When modeling related tasks in computer vision, Multi-Task Learning (MTL) can outperform Single-Task Learning (STL) due to its ability to capture intrinsic relatedness among tasks. However, MTL may encounter the insufficient training problem, i.e., some tasks in MTL may encounter non-optimal situation compared with STL. A series of studies point out that too much gradient noise would lead to performance degradation in STL, however, in the MTL scenario, Inter-Task Gradient Noise (ITGN) is an additional source of gradient noise for each task, which can also affect the optimization process. In this paper, we point out ITGN as a key factor leading to the insufficient training problem. We define the Gradient-to-Noise Ratio (GNR) to measure the relative magnitude of gradient noise and design the MaxGNR algorithm to alleviate the ITGN interference of each task by maximizing the GNR of each task. We carefully evaluate our MaxGNR algorithm on two standard image MTL datasets: NYUv2 and Cityscapes. The results show that our algorithm outperforms the baselines under identical experimental conditions. 

\end{abstract}

\section{Introduction}

\renewcommand{\thefootnote}{\fnsymbol{footnote}}
\footnotetext[1]{Corresponding author. }
\footnotetext[4]{These authors contributed equally. }
\renewcommand{\thefootnote}{\arabic{footnote}}



\noindent Deep learning \cite{DBLP:books/daglib/0040158} has achieved significant success in Multi-Task Learning (MTL) in the field of computer vision. The multi-task model captures the intrinsic correlation among tasks and also allows multiple inferences in one single forward pass, so MTL could achieve the unity of high performance and efficiency in the optimization process \cite{vandenhende2021multi}. However, MTL may suffer from the insufficient training problem \cite{liu2021towards}, which means some tasks in MTL are not optimal compared with the single-task solution. 

Most deep learning models are trained to be optimum by the Stochastic Gradient Descent (SGD) technique \cite{bottou1991stochastic}: models are optimized by the loss gradient based on a mini-batch randomly selected from all data. The gradient obtained by the SGD algorithm has noise \cite{bottou2018optimization,wu2020on}, and many studies \cite{DBLP:conf/iclr/KeskarMNST17,DBLP:conf/nips/HofferHS17} point out that the gradient noise magnitude could affect the SGD generalization in Single-Task Learning (STL). This viewpoint gives us a key insight: in the MTL scenario, due to the joint learning of multiple tasks, the Inter-Task Gradient Noise (ITGN) could also interfere with the optimization of specific tasks, which may be the cause of the insufficient training problem. 

\begin{table}[t]
    \centering
    \caption{Percentage change in performance\protect\footnotemark[1] of MTL relative to STL. NYUv2 and CityScapes are two one-to-many image datasets, where \textbf{Seg.}, \textbf{Dep.}, and \textbf{SN.} denote Segmentation task, Depth task, and Surface Normal task, respectively.}
    \begin{threeparttable}
    \begin{tabular}{lccccc}
    \toprule
    \multirow{2}{*}{\bf Task}  & \multicolumn{3}{c}{\bf NYUv2} & \multicolumn{2}{c}{\bf CityScapes} \cr
    \cmidrule(lr){2-4}\cmidrule(lr){5-6} & \bf Seg. & \bf Dep. & \bf SN. & \bf Seg. & \bf Dep. \cr 
    
    
    \midrule
        performance & \quad 8.3\% $\uparrow$ \quad & \quad 6.9\% $\uparrow$ \quad & \quad 9.7\% $\downarrow$ \quad & \quad 1.1\% $\uparrow$ \quad & \quad 13.2\% $\downarrow$ \quad \cr
    \bottomrule
    \end{tabular}
    \end{threeparttable}
    \label{tab1}
\end{table}
\noindent\footnotetext[1]{Compared STL and MTL at the same settings. Details in Section \ref{c5}. }

\begin{figure}[t]
    \centering
    \subfigure[NYUv2]{
    \includegraphics[width=0.37\columnwidth]{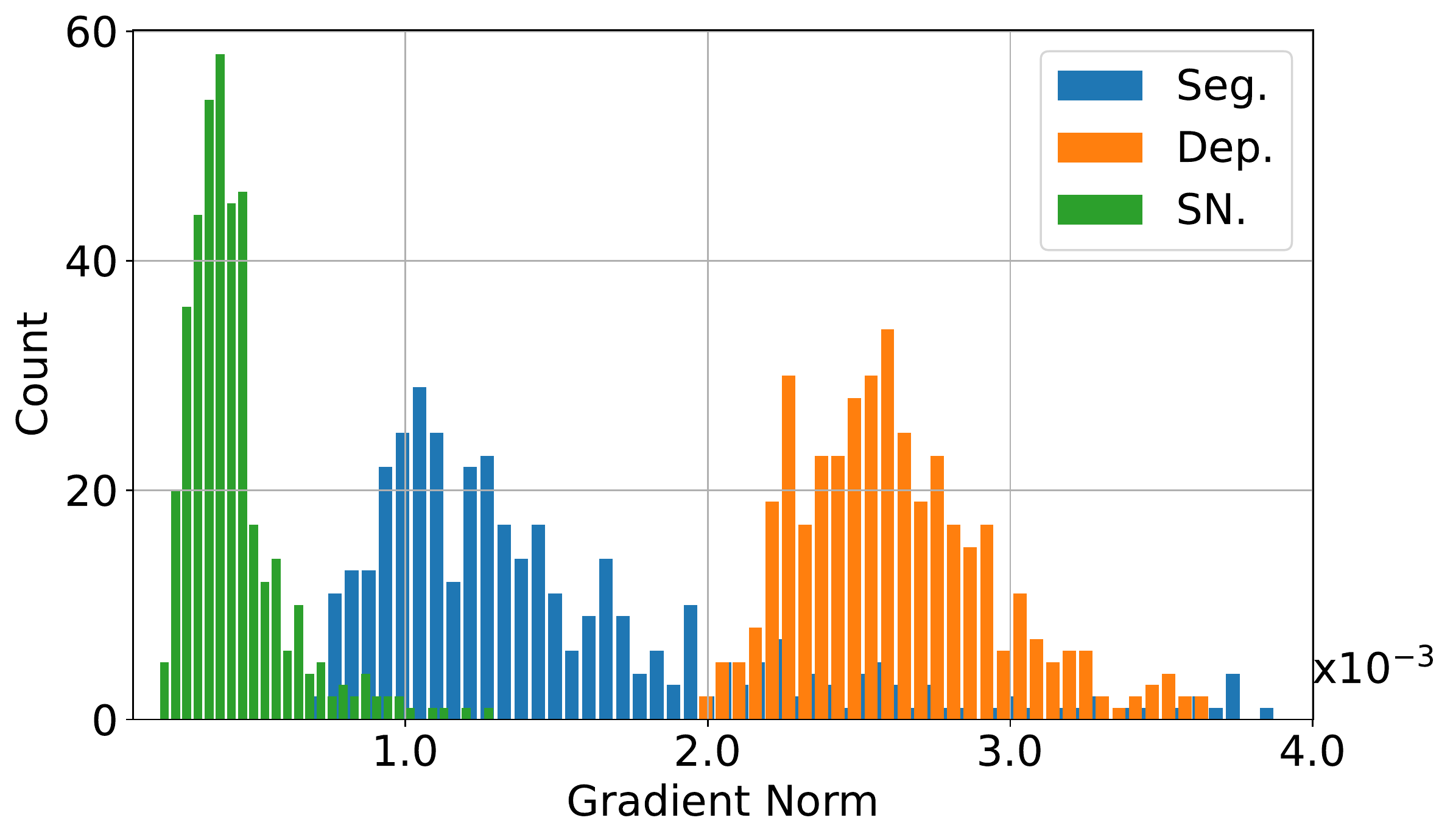} 
    \label{f1-1a}
    }
    \subfigure[CityScapes]{
    \includegraphics[width=0.37\columnwidth]{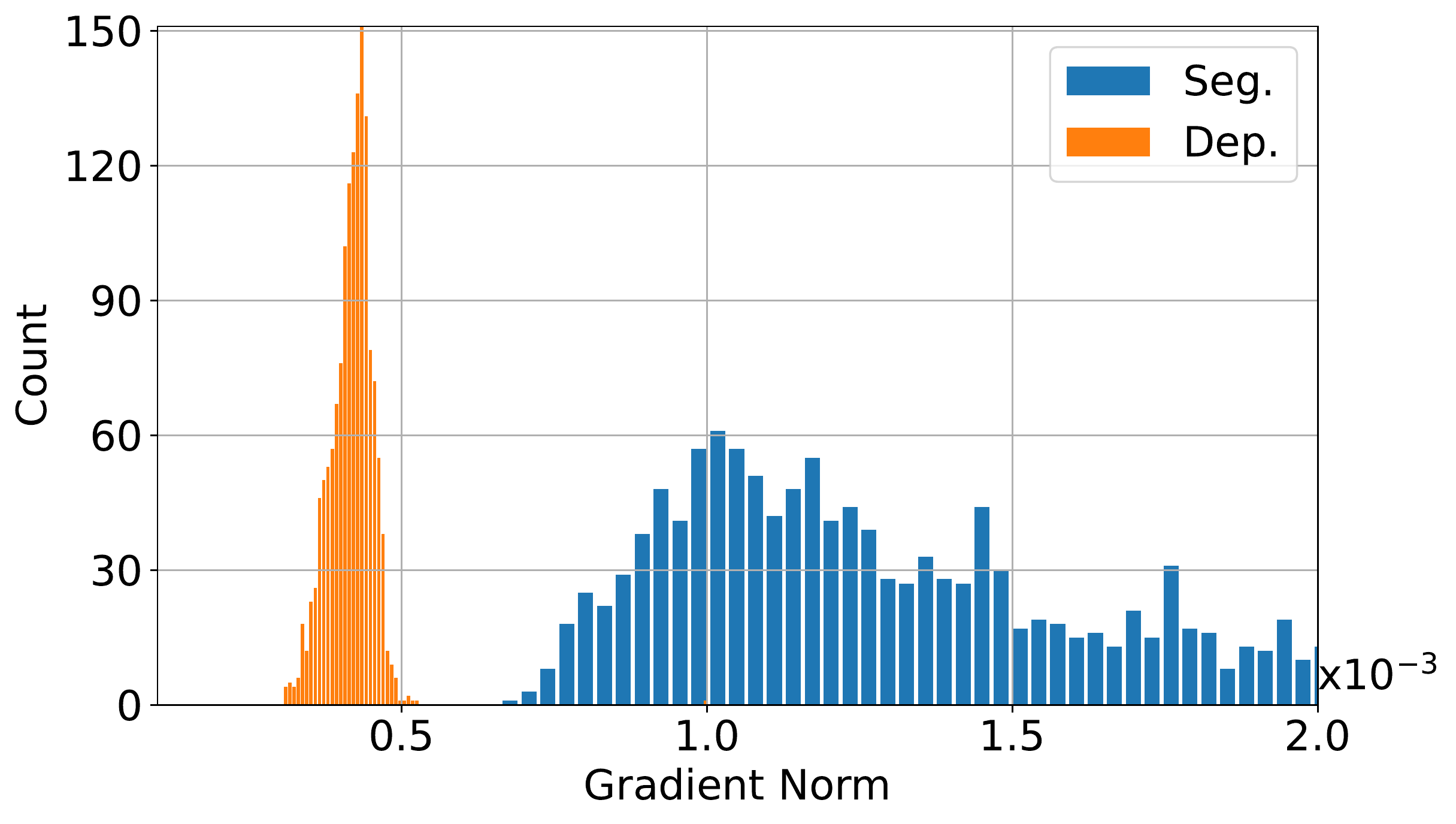}
    \label{f1-1b}
    }
    \caption{Gradient Norm distribution on NYUv2 and CityScapes. } 
\label{f1-1}
\end{figure}

Following this insight, we analyze the performance of each task (Tab. \ref{tab1}) as well as gradient norm distribution (Fig. \ref{f1-1}) on two image MTL datasets. We find that the variance of the gradient norm varies widely across tasks, which implies that different tasks typically have widely varying gradient noise. At the same time, small gradient tasks tend to suffer from performance deterioration in MTL, which shows the relative disadvantage of small gradient tasks in MTL. In this paper, we demonstrate that Inter-Task Gradient Noise (ITGN) is one of the key optimization challenges in MTL (Section \ref{c3}). In fact, the gradient in MTL optimization contains the noises of all tasks, so ITGN is an additional source of gradient noise compared to STL. As a result, some tasks (especially those with small gradients) may suffer from performance deterioration due to the effect of ITGN because the model has difficulty converging to the optimal position based on the gradient when the gradient noise is too high. 

To alleviate the ITGN effect in MTL, we design a dynamic weight strategy called MaxGNR. Specifically, we realize that the magnitude of gradients and noise may vary greatly from task to task, and to quantitatively describe the effect of noise on SGD optimization, we define the Gradient-to-Noise Ratio (\text{GNR}) to measure the relative magnitude of gradient noise. Further, we explain that maximizing the GNR of each task in the MTL scenario is a reasonable method to mitigate ITGN interference (Section \ref{c4-1}). Then, we propose the momentum method to approximate the theoretical GNR to a computable expression (Section \ref{c4-2}), therefore, we can dynamically select the weights that maximize GNR at each iteration (Section \ref{c4-3}). As a result, the ITGN interference is reduced through the MaxGNR algorithm, and the insufficient training problem could be mitigated. 



Furthermore, we extensively examine our \text{MaxGNR} algorithm on two standard image MTL datasets: NYUv2 \cite{silberman2012indoor} and CityScapes \cite{cordts2016the}, and the experimental results reveal that the \text{MaxGNR} algorithm obtain superior performance than other baselines under the same premise of other settings. Our contributions are: 
\begin{itemize}
    \item We analyze the effect of gradient noise on MTL and connect the insufficient training problem with ITGN interference. 
    \item We define the Gradient-to-Noise Ratio (\text{GNR}) to measure the relative magnitude of gradient noise in MTL, and quantitatively describe the ITGN interference by GNR. 
    \item We propose the \text{MaxGNR} algorithm, a dynamic weight strategy to alleviate ITGN interference and experiments demonstrate the algorithm's validity. 
\end{itemize}



\section{Related Work}

\subsection{Multi-Task Learning}

There are mainly two aspects in the field of MTL: network architecture improvement \cite{zhang2018joint,vandenhende2020mti} and optimization strategy development, aiming at feature extraction and balance of performance in MTL, respectively. There have been lots of studies on multi-task architecture improvements, some works focus on the encoder structure \cite{misra2016cross,ruder2019latent,gao2019nddr,liu2019end} and others focus on the decoder part \cite{xu2018pad,zhang2018joint,vandenhende2020mti}. In this paper, we focus on optimization strategy development. A major challenge of MTL is how to balance the performance of tasks in joint learning \cite{vandenhende2021multi}. To solve this challenge, researchers have proposed many algorithms, here we divide these algorithms into two categories: \emph{coarse-grained} and \emph{fine-grained}. The coarse-grained algorithm is to design optimization strategy according to the metrics of prediction, Uncertainty \cite{cipolla2018multi} utilizes homeostatic uncertainty to balance each loss; Dynamic Weight Averaging \cite{liu2019end} tries to balance the decline rate of different task losses; Dynamic Task Prioritization \cite{guo2018dynamic} focuses on the balance of key performance indicators for multiple tasks; GLS \cite{chennupati2019multinet} uses the geometric mean of task-specific losses as the target loss. The fine-grained algorithm is based on the gradient of model parameters, which can more accurately reflect the updating direction in each iteration. In recent years, such algorithms have been developed rapidly. GradNorm \cite{chen2018gradnorm} stimulates the task-specific gradients to be of similar magnitude; MGDA \cite{sener2018multi} regards MTL as multi-objective optimization; PCGrad \cite{yu2020gradient} cut the conflict gradient of different tasks. However, although gradient noise is an important problem in STL, the effect of gradient noise in MTL has never been considered. 

\subsection{Stochastic Gradient Descent} 

Stochastic Gradient Descent (SGD) is one of the standard methods \cite{bottou1991stochastic} to optimize machine learning models. It is originally proposed to make up for the computational bottleneck of gradient descent (GD). Some studies focus on the generation capability of SGD. \cite{zhu2019the} reports SGD outperforms GD, \cite{DBLP:conf/nips/HofferHS17,DBLP:conf/iclr/KeskarMNST17} deduce theoretically that small-batch SGD generalizes better than large-batch SGD. The gradient obtained by the SGD algorithm has un-biased gradient noise \cite{mandt2017stochastic}. To suppress the optimization oscillation caused by gradient noise, \cite{sutskever2013on} proposes to introduce the momentum method in the optimization process, and many optimizers absorb the idea of momentum method, such as Adagrad \cite{duchi2011adaptive}, Adadelta \cite{zeiler2012adadelta}, Adam \cite{kingma2015adam}. 

\section{Effect of Gradient Noise}
\label{c3}

In this section, we analyze the source of gradient noise from the perspective of SGD (Section \ref{c3-1}), and analyze the effect of gradient noise on STL (Section \ref{c3-2}) and MTL (Section \ref{c3-3}). 

\subsection{Preliminaries of SGD}
\label{c3-1}

Formally, the design of the machine learning algorithm is based on data $D = \left\{ \left( X_i,Y_i \right) \right\}$, the hypothesis function $F$, and loss function $l$. Algorithms are designed to search the optimal parameter $\{F_\theta \vert \theta \in \Theta \subset \mathbb{R}^d\}$ to get the lowest expected risk $\mathcal{R}(\theta)$ under the loss function $l$, where $\theta$ is the parameter of the hypothesis function and $d$ is the dimension of the parameter. The expected risk $\mathcal{R}(\theta)$ cannot be obtained, so stochastic Gradient Descent algorithm (SGD) \cite{bottou1991stochastic} selects a mini-batch $S= \left\{ \left( X_i,Y_i \right) \right\}_{\left| S \right|}$ independent and identically distributed (i.i.d.) from the data $D$, and the expected risk $\mathcal{R}(\theta)$ can be estimated by the empirical risk $\widehat{\mathcal{R}}(\theta)$ calculated by mini-batch. In this paper, $\nabla_{\theta}\mathcal{R}(\theta)$ and $\nabla_{\theta}\widehat{\mathcal{R}}(\theta)$ are denoted as $g(\theta)$ and $\widehat{g_{S}}(\theta)$, and $l(F_{\theta}(X_i),Y_i)$ is expressed as $l_i(\theta)$. 



The gradient $\widehat{g_{S}}(\theta)$ obtained by SGD is stochastic, so the gradient can be decomposed into the expected gradient and gradient noise. Many works \cite{mandt2017stochastic,DBLP:conf/nips/HeLT19} assume that the gradient noise belongs to the Gaussian class due to the classical central limit theorem \cite{mandt2017stochastic}. In this paper, we follow this assumption in order to describe the effect of gradient noise on MTL quantitatively. 

\begin{theo}
\label{theo1}
An individual data $\left( X_i,Y_i \right)$ is selected independent and identically distributed (i.i.d.) from data $D$, and its gradient $\nabla_{\theta} l_i(\theta)$ obeys the Gaussian distribution: 
\begin{equation}
    \nabla_{\theta} l_i(\theta) \sim \mathcal{N}(g(\theta),C)
    \label{e3-2}
\end{equation}


\noindent where $g(\theta)$ is the expected gradient, $C$ is the covariance matrix, and is approximately constant for $\theta$, which is determined by data $D$ and loss function $l$. 
\end{theo}

\subsection{Gradient Noise in STL} 
\label{c3-2}

Many studies \cite{DBLP:conf/iclr/KeskarMNST17,DBLP:conf/nips/HofferHS17} emphasize the magnitude of gradient noise could effect the generalization of SGD. In practice, noise magnitude can be easily adjusted by batch size $\left| S \right|$ \cite{DBLP:conf/nips/HeLT19}. According to Assumption \ref{theo1}, $\widehat{g_{S}}(\theta)$ can be simplified as: 
\begin{equation}
\begin{split}
    \widehat{g_{S}}(\theta) =& \frac{1}{\left| S \right|}\sum_{i=1}^{\left| S \right|} \nabla_{\theta} l_i(\theta) =  g(\theta) + n_{g(\theta)}\\
    & \text{where} \quad n_{g(\theta)} \sim \mathcal{N}(0,\frac{C}{\left| S \right|})
\end{split}
\label{e3-3}
\end{equation}

\noindent where $n_{g(\theta)}$ is the gradient noise. Because of the Positive Semi-definite of $C$, the noise magnitude can be measured as: 
\begin{equation}
    \mathbb{E}[\| n_{g(\theta)} \|^2] = tr {(C)}/\left| S \right|
    \label{e3-4}
\end{equation}

\begin{figure}[t]
    \centering
    \subfigure[Perf. vs. Batch Size]{
    \includegraphics[width=0.37\columnwidth]{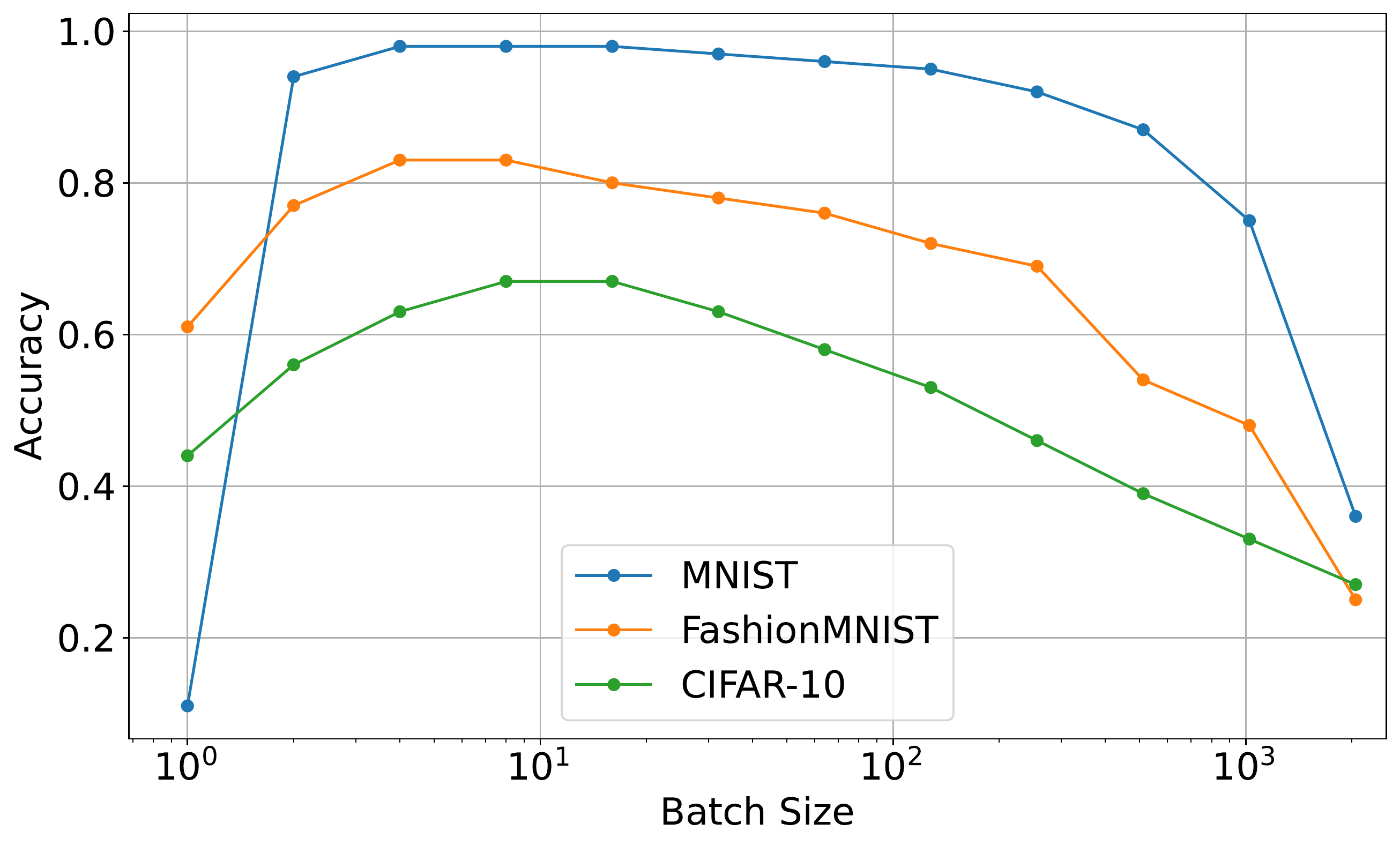} 
    \label{f3-1a}
    }
    \subfigure[Perf. vs. Noise]{
    \includegraphics[width=0.37\columnwidth]{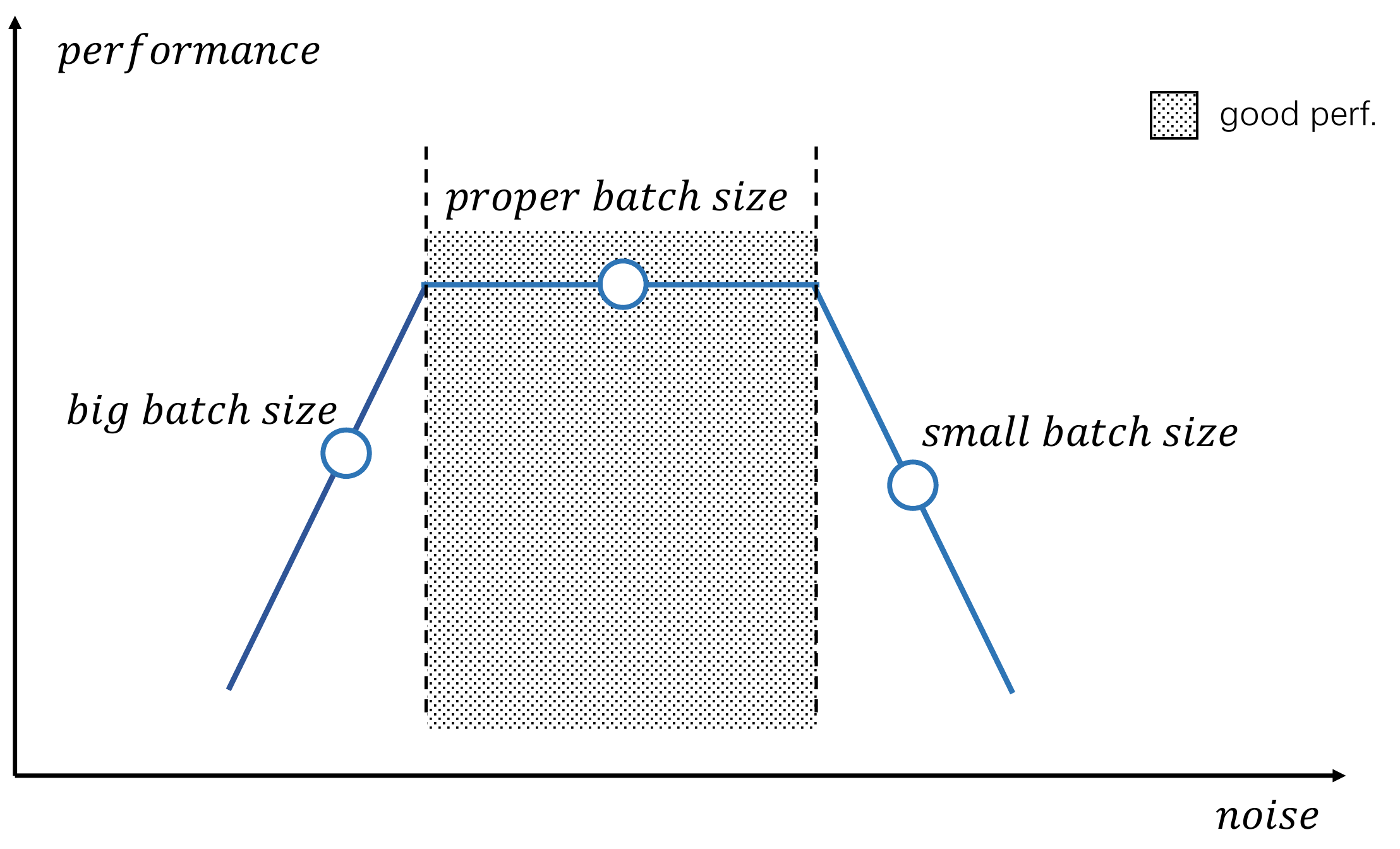}
    \label{f3-1b}
    }
    \caption{The effect of gradient noise magnitude on the performance of STL. The noise magnitude is controlled by the batch size. We conduct experiments on MNIST, FashionMNIST, and CIFAR10. As the batch size increases, performance first increases, then stabilizes, and finally decreases on all three datasets. }
    \label{f3-1}
\end{figure}

By adjusting the batch size $\left| S \right|$, we evaluated the impact of noise magnitude on model performance. The result in Fig. \ref{f3-1a} showed that the performance can be maintained only in the appropriate noise magnitude range. In fact, large noise would interfere with the convergence of the model, while the model is unable to escape the local optimum with small noise, so the effect of gradient noise on performance can be expressed as Fig. \ref{f3-1b}. 

\subsection{Inter-Task Gradient Noise}
\label{c3-3}

In the multi-task SGD process, multiple tasks have their own loss functions $\left\{ l^k \right\}$, and Weighted Average Method (WAM) is a mainstream method to combine multiple loss functions. Specifically, WAM can be expressed as: 
\begin{equation}
    \begin{split}
        L=\sum_{k=1}^n \omega_k l^k(\theta) \quad s.t. \sum_{k=1}^n \omega_k = 1
    \end{split}
    \label{e3-5}
\end{equation}

According to Assumption \ref{theo1}, the SGD gradient in MTL is expressed as follows: 
\begin{equation}
    \begin{split}
    \widehat{g_{S}}(\theta) &= \sum_{k=1}^n \omega_k g^k(\theta) + \sum_{k=1}^n \omega_k n_{g^k(\theta)}\\
    \end{split}
    \label{e3-6}
\end{equation}

Based on Eq. \ref{e3-6}, ITGN also contributes to task $i$'s optimization in the MTL scenario, where $\text{ITGN}_i$ is as follows: 
\begin{equation}
    \text{ITGN}_i = \sum_{k \neq i} \omega_k n_{g^k(\theta)} \sim \mathcal{N}(0,\frac{\sum_{k \neq i} \omega_k^2 C^k}{\left| S \right|})
    \label{e3-7}
\end{equation}

From Eq. \ref{e3-4}, we can infer that $\left\{ C^k \right\}$ and $\left| S \right|$ control the magnitude of $\left\{ n_{g^k(\theta)} \right\}$, although multiple tasks share $\left| S \right|$, $\left\{ C^k \right\}$ are task-specific and determined by the intrinsic characteristics of tasks, so $\left\{ n_{g^k(\theta)} \right\}$ usually exist huge discrepancies, as shown in Fig. \ref{f1-1}. Therefore, ITGN can cause performance deterioration in MTL (similar to the performance deterioration due to large noise in STL). 

\begin{figure*}[t]
\centering
\subfigure[Performance in STL]{
\includegraphics[width=0.305\columnwidth]{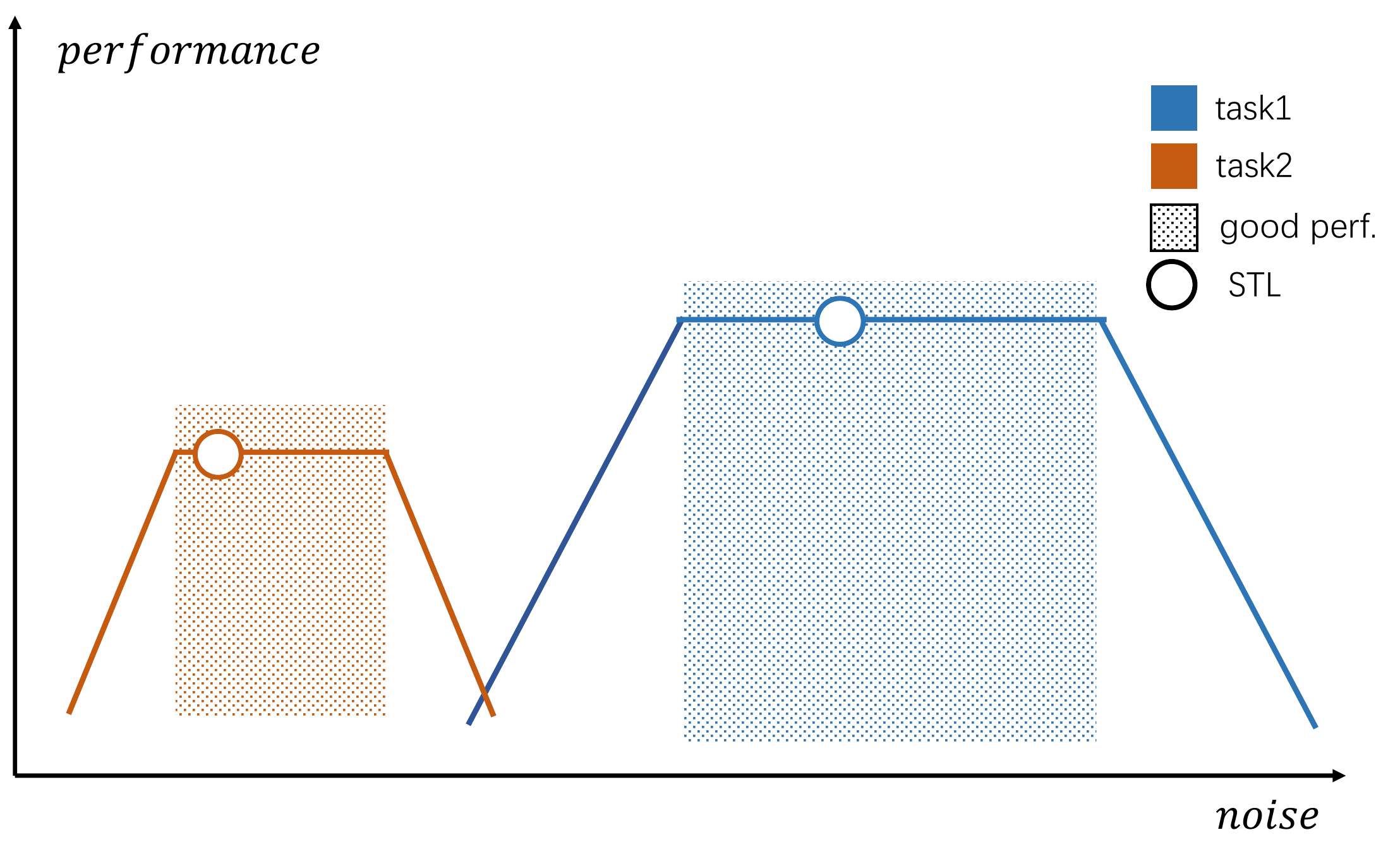} 
\label{f4-1a}
}
\subfigure[Performance in MTL]{
\includegraphics[width=0.305\columnwidth]{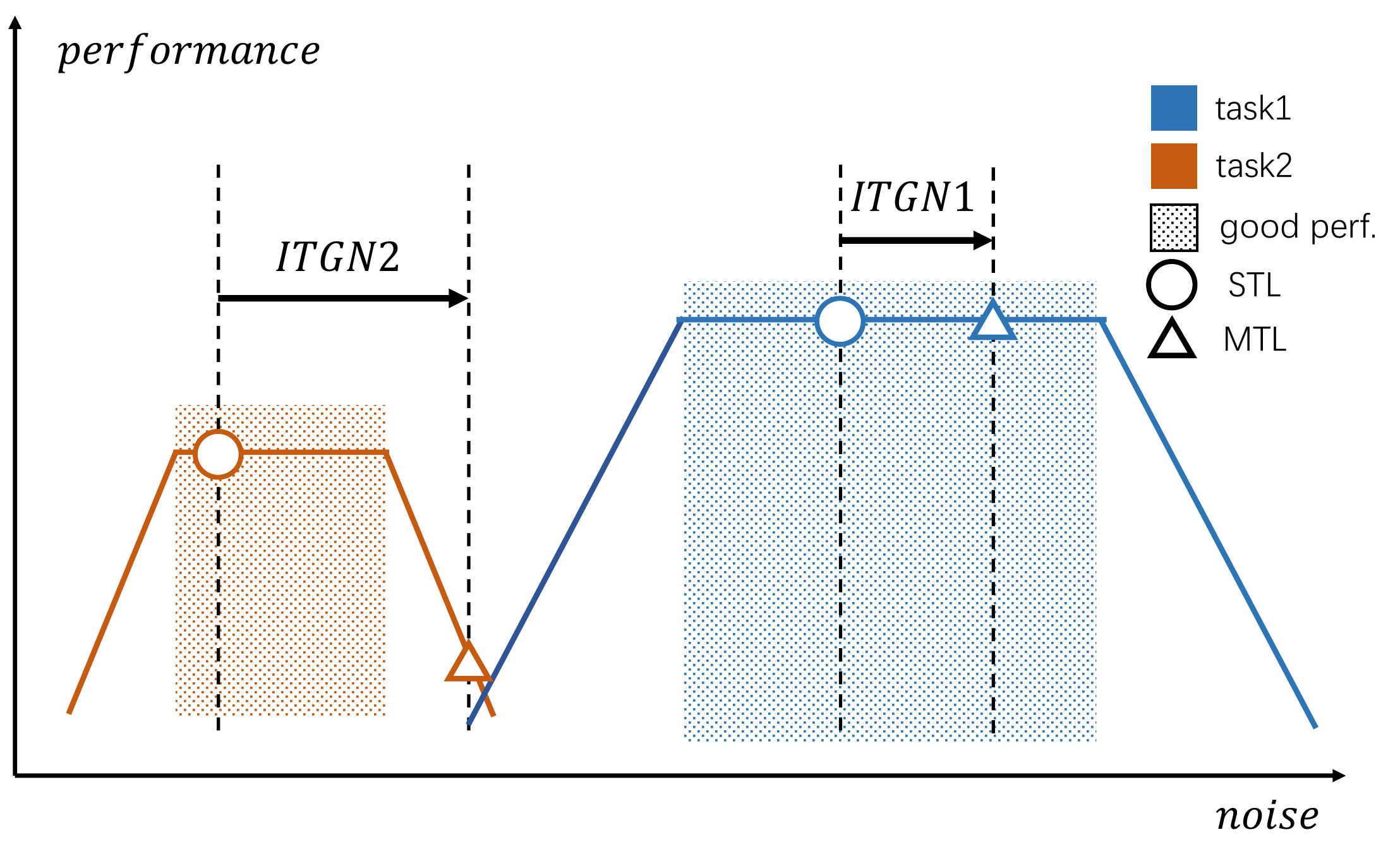} 
\label{f4-1b}
}
\subfigure[Design weights to maximize GNR]{
\includegraphics[width=0.305\columnwidth]{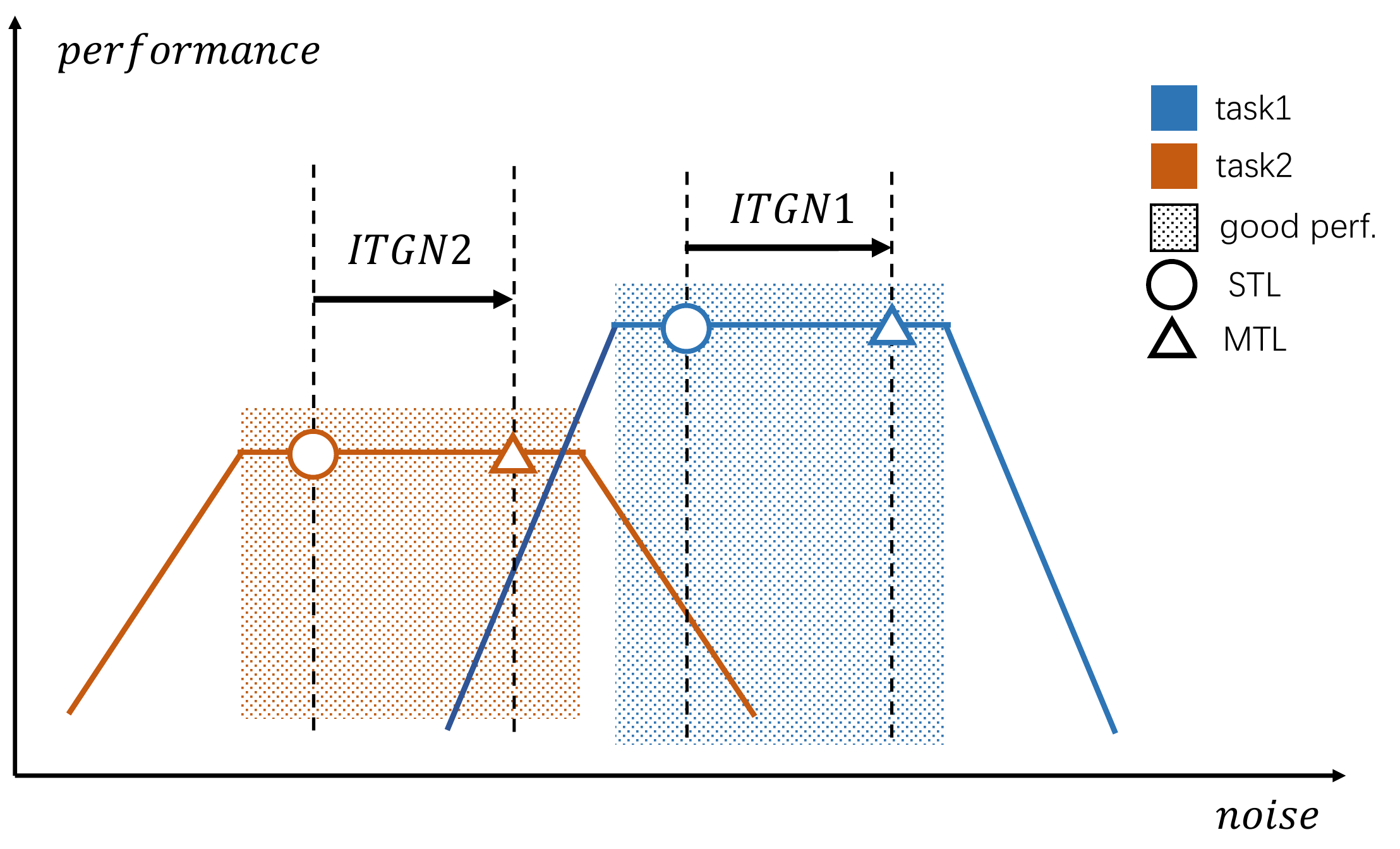} 
\label{f4-1c}
}
\caption{Illustration of MaxGNR algorithm. In the STL scenario, reasonable settings (model, batch size, learning rate, etc.) can be selected to ensure that the gradient noise is in the appropriate range, as in Fig. \ref{f4-1a}. In the MTL scenario, ITGN is also a source of gradient noise that can affect the optimization of specific tasks. Therefore, ITGN may cause task-specific performance deterioration due to the large differences in ITGN, as is illustrated in Fig. \ref{f4-1b}. Our MaxGNR algorithm attempts to alleviate the ITGN interference as Fig. \ref{f4-1c}. According to Eq. \ref{e4-5}, We can design appropriate weights to maximize GNR for the purpose of mitigating ITGN interference. }


\label{f4-1}
\end{figure*}

\begin{algorithm}[t]
\setstretch{1.35}
\SetAlgoLined
\KwIn{data $\{X_i,Y_i^1,\dots,Y_i^n\}_{|D|}$, learning rate $\mu$. } 
\KwOut{optimal network parameters $\theta$. }

Initialize network parameters $\theta_0$ \;
  
\For{t = 0 to T}{
select mini-batch data $S=\{X_i,Y_i^1,\dots,Y_i^n\}_{|S|}$\;

compute each empirical risk $\{ \widehat{\mathcal{R}^k}(\theta_t) \}$\;

compute the gradient of empirical risk $\{\nabla \widehat{\mathcal{R}^k}(\theta_t)\}$\;


compute the momentum gradient $\{ m_t^k \}$ and noise $\{n_t^k \}$ \;

select $\{ \omega_t^k \}$ by $\argmax_{\{\omega_k\}}\left\{ \min \{ \text{GNR}_k ( m_t^k, \{n_t^k \} )\} \right\}$ \;


update network parameters $\theta_{t+1} = \theta_{t} - \mu \nabla_{\theta_t}\sum_{k=1}^n \omega_t^k \widehat{\mathcal{R}^k}(\theta_t) $ \;
}
    \caption{\text{MaxGNR} algorithm}
    \label{algorithm1}
\end{algorithm}


\section{Method}
\label{c4}

We describe the necessity of controlling the noise magnitude range to maintain excellent model performance in Section \ref{c3-2} as Fig. \ref{f4-1a} and explain that ITGN could also interfere with the specific task's optimization in MTL in Section \ref{c3-3}. In summary, the range of ITGN may vary greatly, so it can be inferred that ITGN is likely to cause performance deterioration on some tasks as Fig. \ref{f4-1b}. In this section, we propose the \text{MaxGNR} algorithm, a dynamic weight strategy to minimize the ITGN interference in MTL as Fig. \ref{f4-1c}. 

Gradient-to-Noise Ratio (GNR) is the core concept of this paper, which is defined to measure the relative magnitude of gradient noise (Section \ref{c4-1}), and it can be found that ITGN would reduce GNR compared to STL because ITGN is an additional gradient noise source. So it is a viable method to mitigate ITGN interference by maximizing GNR. To solve the dilemma that GNR is not computable, we propose to estimate $\{ g^k(\theta) \}$ and $\left\{ n_{g^k(\theta)} \right\}$ using the momentum method, and thus estimate a computable GNR (Section \ref{c4-2}). Eventually, we can dynamically select the weights that maximize GNR in each iteration, thus achieving the purpose of alleviating ITGN interference (Section \ref{c4-3}). The whole process of the MaxGNR algorithm is shown in Alg. \ref{algorithm1}. 

\subsection{Gradient-to-Noise Ratio}
\label{c4-1}


In order to measure the noise interference on stochastic gradient optimization, it is necessary to define an appropriate metric to measure the relative magnitude of gradient noise to the gradient. 


\begin{defi}
In the process of SGD, for a specific task, the expected gradient is $g(\theta)$, the random variable that could make $\widehat{g_{S}}(\theta)$ deviate from $g(\theta)$ is the gradient noise $n_{g(\theta)}$, we define \textbf{Gradient-to-Noise Ratio (GNR)} as: 
\begin{equation}
    \text{GNR} = \frac{ \| g(\theta) \|^2}{\mathbb{E}[\| n_{g(\theta)} \|^2] }
\end{equation}
\label{d1}
\end{defi}

GNR reflects the relative magnitude of gradient noise to the gradient. The homogeneity ensures that GNR is independent of weight in STL, and according to Eq. \ref{e3-4}, $\mathbb{E}[\| n_{g(\theta)} \|^2]$ is an approximate constant and easy to represent.

There is only one source of gradient noise in STL. According to definition \ref{d1}, $\text{GNR}_S$ is represented as: 
\begin{equation}
    \text{GNR}_S = \frac{ \| g(\theta) \|^2}{\mathbb{E}[\| n_{g(\theta)} \|^2] } = \frac{ \| g(\theta) \|^2}{tr {(C)}/\left| S \right| }
    \label{e4-1}
\end{equation}

In the MTL scenario, due to the ITGN interference, task $k$'s GNR is represented as: 
\begin{equation}
    \text{GNR}_M^k = \frac{ \| \omega_k g^k(\theta) \|^2}{\mathbb{E}[\| \omega_k n_{g^k(\theta)} + \text{ITGN}_k \|^2]}
    \label{e4-2}
\end{equation}


According to Eq. \ref{e3-4} and Eq. \ref{e3-6}, $\mathbb{E}[ \| \text{ITGN}_k \|^2]$ could be represented by $\{ C^k \}$ and $| S |$, and due to the linear property of matrix’s trace: 
\begin{equation}
    tr(m \cdot A + n \cdot B)=m \cdot tr(A)+n \cdot tr(B)
    \label{e4-3}
\end{equation}

$\text{GNR}_M^k$ is eventually simplified as: 
\begin{equation}
\begin{split}
    \text{GNR}_M^k = \frac{ \| \omega_k g^k(\theta) \|^2}{\sum_{k=1}^n \omega_k^2 tr (C^k) /\left| S \right| } \leq \text{GNR}_S^k
\end{split}
\label{e4-4}
\end{equation}

In general, hyperparameters with superior performance in STL would be selected for MTL, so we assume that $\text{GNR}_S^k$ can be controlled as an appropriate GNR. However, ITGN interference reduces the GNR of each task, if $\text{ITGN}_k \gg n_{g^k(\theta)}$, $\text{GNR}_M^k$ could be much smaller than $\text{GNR}_S^k$, which is likely to cause performance deterioration of task $k$. Therefore, selecting weights $\{\omega_k\}$ to maximize $\text{GNR}_M^k$ is a feasible method to alleviate ITGN interference for task $k$. In the MTL scenario, to ensure that all tasks could alleviate ITGN interference as much as possible, we should maximize $\min \{ \text{GNR}_M^k \}$ as: 
\begin{equation}
    \{\omega_k\} = \argmax_{\{\omega_k\}} \left\{ \frac{\min \left\{ \Vert \omega_k g^k(\theta)\Vert^2 \right\} }{\sum_{k=1}^n \omega_k^2 tr (C^k) /\left| S \right|} \right\}
   \label{e4-5}
\end{equation}

\subsection{Estimation of Expected Gradient}
\label{c4-2}


However, GNR is a theoretical concept and it is not realistic to calculate $\left\{ g^k(\theta_t) \right\}$ and $\left\{ C^k \right\}$, therefore, we should design a method to estimate them. During the training process, we can only get the empirical gradient $\{ \widehat{g_{S}}^k(\theta_t) \}$. In this paper, we propose that the momentum method can be used to estimate the expected gradient and gradient noise. 

\noindent \textbf{Momentum Method} \quad The momentum method is widely applied in various optimizers. AdaGrad, AdaDelta, Adam, and so on adopt the momentum method as a technique to stabilize gradient. Momentum is expressed as $m$, and we set $m_0 = \widehat{g_{S}}(\theta_0)$. The basic process of the momentum method is as follows: 
\begin{equation}
  \begin{split}
    &m_t=\gamma \cdot m_{t-1}+(1-\gamma) \cdot \widehat{g_{S}}(\theta_t)\\
    &\theta_t = \theta_{t-1} - \mu \cdot m_t
  \end{split}
   \label{e4-6}
\end{equation}

\noindent where $\gamma$ is the decay rate, $\mu$ is the learning rate, both are hyperparameters. It is an option to estimate the expected gradient by momentum gradient, because the momentum gradient could reduce noise and stabilize gradient. According to Eq. \ref{e3-3} and Eq. \ref{e4-6}, the momentum gradient can be derived as: 
\begin{equation}
  \begin{split}
    m_t^k &\approx (1-\gamma)\sum_{i=1}^t \gamma^{t-i} \cdot g^k(\theta_i) + n_{m_t^k}\\
    &\text{where} \quad n_{m_t^k} \sim \sqrt{\frac{1-\gamma}{1+\gamma}} \mathcal{N}(0,\frac{C}{\left| S \right|})
  \end{split}
   \label{e4-7}
\end{equation}

By comparing Eq. \ref{e3-3} and Eq. \ref{e4-7}, it can be found that the momentum method produces a contraction coefficient of $\sqrt{(1-\gamma) / (1+\gamma)}$ on noise magnitude, the larger $\gamma$ is, the smaller the estimated noise magnitude is. At the same time, the cost is that the momentum gradient $m_t^k$ is no longer the un-biased estimation of $g^k(\theta_t)$, the larger $\gamma$ is, the greater the estimation error is. So $\gamma$ becomes a trade-off factor of the expected gradient estimation, if the estimation error of $g^k(\theta_t)$ is too large, the momentum method is also invalid. Fortunately, parameter optimization is a long process, and the change of parameters in each iteration is small compared to the parameters themselves. Based on this fact, we make the following assumption: 

\begin{theo}
\label{theo2}
the expected gradient changes slowly and the expected gradients in adjacent iterations are similar: 
\begin{equation}
    g^k(\theta_n) \approx g^k(\theta_{n + \Delta n}), \ \text{when $\Delta n$ is small}
\end{equation}
\end{theo}

On the premise of Assumption \ref{theo2}, the estimation error of $g^k(\theta_t)$ in Eq. \ref{e4-7} can be reduced to a large extent, so it is reasonable to estimate $g^k(\theta_t)$ with $m^k(\theta_t)$. 

\subsection{Weight Selection}
\label{c4-3}

According to the estimation method proposed by Eq. \ref{e4-7}, the empirical gradient of each task can be decomposed into momentum gradient $m^k(\theta)$ and gradient noise $n_{g^k}(\theta)$. Therefore, we can estimate $n_{g^k}(\theta)$: 
\begin{equation}
\begin{split}
    n_{g^k}(\theta) = \widehat{g_{S}}^k(\theta) - m^k(\theta) \\
\end{split}
\label{e4-8}
\end{equation}

To calculate the GNR of each task, we can replace $g^k(\theta)$ with $m^k(\theta)$ according to Eq. \ref{e4-7} and Assumption \ref{theo2}, and although $\mathbb{E}[\| n_{g^k(\theta)} \|^2]$ is approximately determined, each gradient noise is a random variable, so it is reasonable to use $\| \sum_{k=1}^n \omega_k n_{g^k(\theta)} \|^2$ to replace $\sum_{k=1}^n \omega_k^2 tr (C^k) /\left| S \right|$. Therefore, the \text{MaxGNR} algorithm can ultimately expressed as: 
\begin{equation}
    \{\omega_k\} = \argmax_{\{\omega_k\}} \left\{ \frac{\min \left\{ \Vert \omega_k m^k(\theta)\Vert^2 \right\} }{\| \sum_{k=1}^n \omega_k n_{g^k(\theta)} \|^2}  \right\}
   \label{e4-9}
\end{equation}

Because of the non-linearity of Eq. \ref{e4-9}, a general analytical solution cannot be found. We use the steepest descent method \cite{DBLP:books/cu/BV2014} to obtain the optimal $\{\omega_k\}$ at each iteration. Compared with the number of parameters in the neural network, the number of parameters in the steepest descent method is the number of tasks, so the computation cost can be ignored. At the same time, the computation cost of the steepest descent method is not sensitive to the number of tasks, which is different from grid search. 


\section{Experiments}
\label{c5}

\begin{table*}[t]
  \centering
  \caption{Experiment results of different algorithms for NYUv2, and we split the table into coarse-grained and fine-grained algorithms. The \textbf{bold} represents the top2 scores.}
  \begin{threeparttable}
    \begin{tabular}{lcccccc}  
    \toprule  
    \multirow{2}{*}{\bf Algorithm} & 
    \multicolumn{2}{c}{\bf Segmentation}&\multicolumn{2}{c}{\bf Depth}&\multicolumn{2}{c}{\bf Surface Normal} \cr 
    \cmidrule(lr){2-3} \cmidrule(lr){4-5} \cmidrule(lr){6-7}  
    & mIoU $\uparrow$ & Pix Acc $\uparrow$ & Abs Err $\downarrow$ & Rel Err $\downarrow$ & Mean $\downarrow$ & Median $\downarrow$ \cr  
    \midrule  
    single task & 26.15 & 53.27 & 0.6655 & 27.89 & 30.57 & 25.12 \cr 
    equal weights & 28.32 & 55.60 & 0.6196 & 27.61 & 31.80 & 26.82 \cr 
    \midrule
    \emph{coarse-grained} \cr
    DWA & 28.43 & 56.08 & 0.6075 & 25.08 & 31.98 & 26.89 \cr 
    Uncertainty & 28.06 & 54.46 & 0.6059 & 25.69 & 31.49 & 26.32 \cr 
    \midrule
    \emph{fine-grained} \cr
    GradNorm & 28.92 & 56.26 & 0.6057 & \bf24.06 & 31.66 & 26.49 \cr 
    MGDA & 22.38 & 49.59 & 0.7414 & 27.51 & 30.50 & \bf23.85 \cr 
    PCGrad & 28.72 & 56.17 & 0.6131 & 27.33 & 32.19 & 27.00 \cr 
    \midrule
    \text{MaxGNR} ($\gamma=0.5$) & \bf29.76 & \bf57.64 & \bf0.5943 & \bf24.35 & \bf29.89 & \bf24.38 \cr 
    \text{MaxGNR} ($\gamma=0.9$) & \bf29.48 & \bf56.81 & \bf0.5904 & 24.91 & \bf30.43 & 24.85  \cr 
    \bottomrule
    \end{tabular}
    \end{threeparttable}
    \label{tab2}
\end{table*}

\subsection{Experimental Settings}

\subsubsection{Datasets}

We focused on one-to-many predictions datasets  $\{X_i,Y_i^1,\dots,Y_i^n\}_{|D|}$  \cite{DBLP:conf/ijcai/ZhangXWJ17} in this paper. We evaluated the proposed \text{MaxGNR} algorithm on two image datasets: NYUv2 \cite{silberman2012indoor} and CityScapes \cite{cordts2016the}. NYUv2 is a challenging indoor scene dataset in various room types (bathrooms, living rooms, studies, etc.), and this dataset has three tasks: 13-class semantic segmentation, depth estimation, and surface normal prediction. Although NYUv2 is a relatively small dataset (795 training, 654 test images), it contains both regression and classification tasks, making it a good candidate for testing the robustness of different types of loss functions. CityScapes is a high-resolution street-view images dataset, we chose two sub-tasks for our experiment: semantic segmentation and depth estimation. Compared to NYUv2, the street-view image contained in CityScapes is relatively simpler, because the viewpoints and lighting conditions are relatively fixed, and the appearance of each object class changes little in shape. 

\subsubsection{Implementation Details}

We implemented our experiments based on the network architecture in \cite{liu2019end}. The network architecture had an encoder-decoder structure that was homogeneous across all tasks, where the encoder extracted the representation and the decoder matched individual tasks. For the NYUv2 and CityScapes datasets, we set the batch size to 2 and 128, respectively, and the learning rate of the Adam optimizer to 1e-4. After training each model for 100/200 epochs, we selected the best model on the training set for testing. In our experiment, we set the learning rate of the steepest descent method to 1e-3, and $\{\omega_k\}$ updated 100 steps in each iteration. 

\subsubsection{Evaluation Metrics}

To compare the baselines and our method, we chose two metrics for each task. Following the settings in \cite{cipolla2018multi,liu2019end,yu2020gradient}, we chose mIoU and Pixel Accuracy for Segmentation task, Absolute and Relative Error for Depth task, and Mean and Median Angle Distance for Surface Normal task. 

\subsection{Baselines}

In addition to equal weights and single-task models, the baselines can be divided into two types: coarse-grained algorithms and fine-grained algorithms. Coarse-grained algorithms included DWA \cite{liu2019end} and Uncertainty \cite{cipolla2018multi}, fine-grained algorithms included GradNorm \cite{chen2018gradnorm}, MGDA \cite{sener2018multi} and PCGrad \cite{yu2020gradient}. 

\subsection{Main Results}


\begin{table}[t]
    \centering
    \caption{Experiment results of different algorithms for CityScapes. }
    \begin{threeparttable}
    \begin{tabular}{lcccc}
    \toprule
    \multirow{2}{*}{\bf Algorithm} & 
    \multicolumn{2}{c}{\bf Segmentation}&\multicolumn{2}{c}{\bf Depth} \cr
    \cmidrule(lr){2-3}\cmidrule(lr){4-5} 
    & mIoU $\uparrow$ & Pix Acc $\uparrow$ $\uparrow$ & Abs Err $\downarrow$ & Rel Err  $\downarrow$ \cr
    \midrule
        single task & 66.91 & 90.03 & 0.0153 & 35.85 \cr
        equal weights & 67.63 & 90.09 & 0.0163 & 57.13 \cr
        \midrule
        GradNorm & 67.20 & 90.76 & 0.0169 & 57.21 \cr
        MGDA & 66.83 & 90.88 & 0.0170 & 47.14 \cr
        PGGrad & \bf67.75 & \bf91.01 & 0.0164 & 54.95 \cr
        \midrule
        \text{MaxGNR} ($\gamma=0.5$) & \bf68.01 & \bf91.07 & \bf0.0162 & \bf44.11 \cr
        \text{MaxGNR} ($\gamma=0.7$) & 67.33 & 90.76 & \bf0.0160 & \bf42.48 \cr
    \bottomrule
    \end{tabular}
    \end{threeparttable}
    \label{tab3}
\end{table}

\noindent \textbf{Results on NYUv2} was shown in Tab. \ref{tab2}. Our \text{MaxGNR} algorithm outperformed the baselines in almost all metrics. When specific tasks were analyzed, most approaches outperformed the single-task model in the Depth task, and most baselines performed marginally better than the single-task model in the Segmentation task, while, in the Surface Normal task, almost all baselines suffered from performance deterioration. It was important to note that MGDA, which outperformed the single-task model in the Surface Normal task, cannot balance the optimization performance on all tasks, resulting in MGDA's bad performance in the other two tasks. Our \text{MaxGNR} algorithm not only outperformed the single-task model and baselines in the Segmentation and Depth tasks, but it also performed well in the Surface Normal task and overcame the insufficient training problem. 

\noindent \textbf{Results on CityScapes} was shown in Tab. \ref{tab3}. In CityScapes, the performance of the single-task model was similar to all MTL algorithms on the Segmentation task (about 91\%), while there was a serious performance deterioration in the Depth task. We analyzed the performance of baselines and our algorithm in two tasks. In the Segmentation task, our algorithm was not significantly better than other baselines, but in the Depth task, our algorithm improved the performance of the model and greatly narrowed the gap with the single-task model. 

\subsection{Dynamic Change of Weights}

\begin{figure}[t]
    \centering
    \subfigure[NYUv2]{
    \includegraphics[width=0.35\columnwidth]{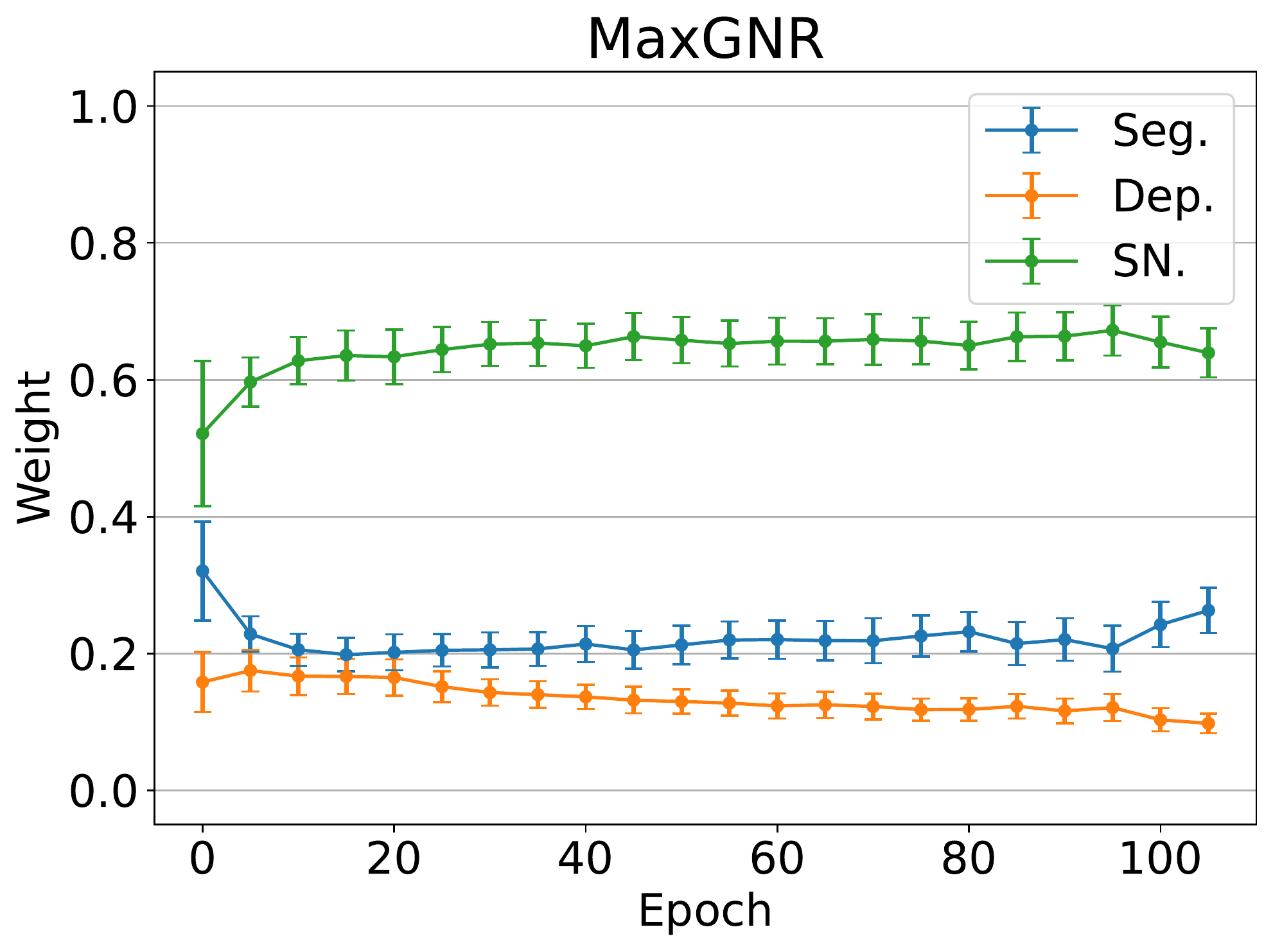}
    \includegraphics[width=0.35\columnwidth]{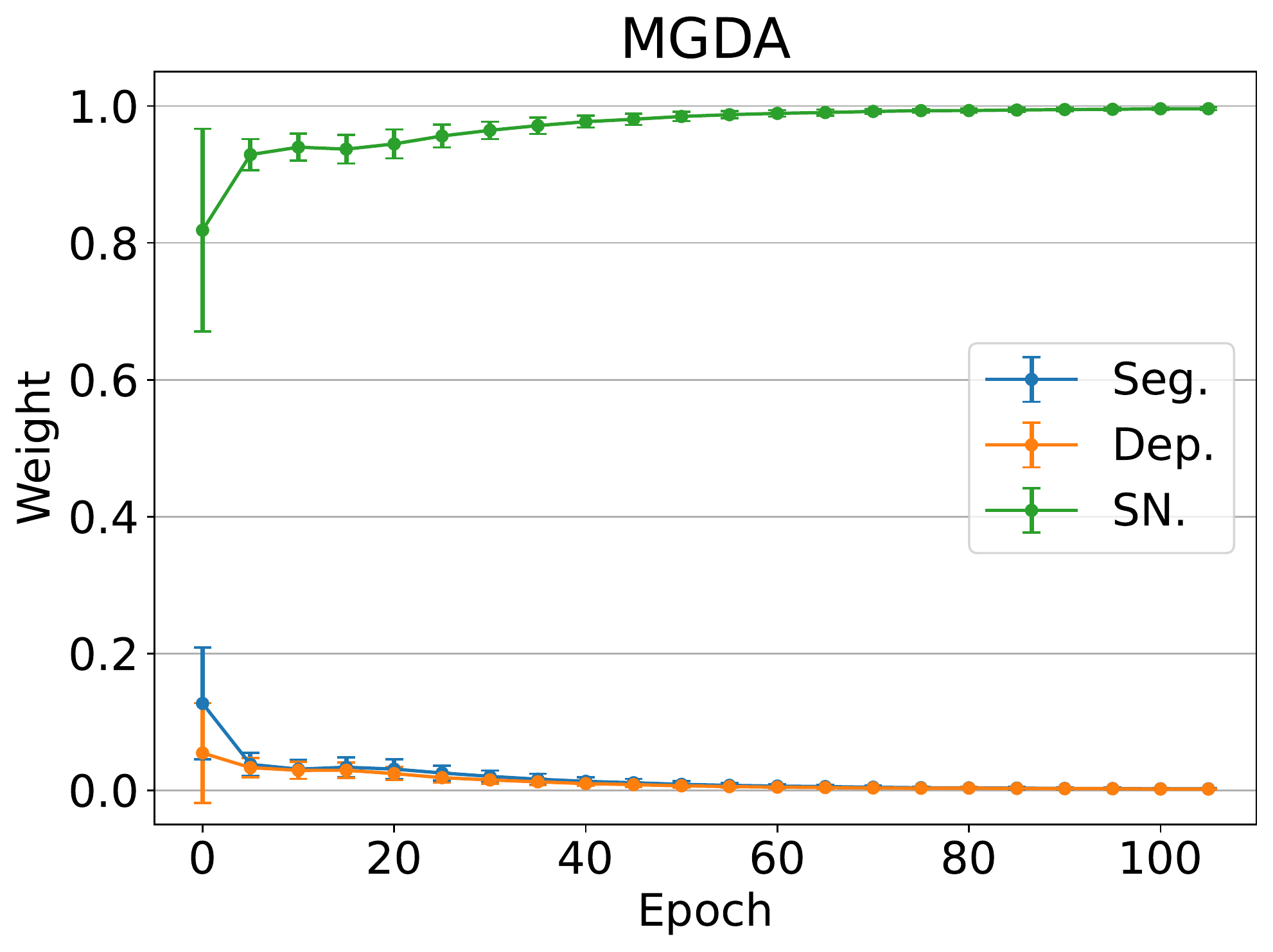}
    \label{f5-2a}
    }
    \subfigure[CityScapes]{
    \includegraphics[width=0.35\columnwidth]{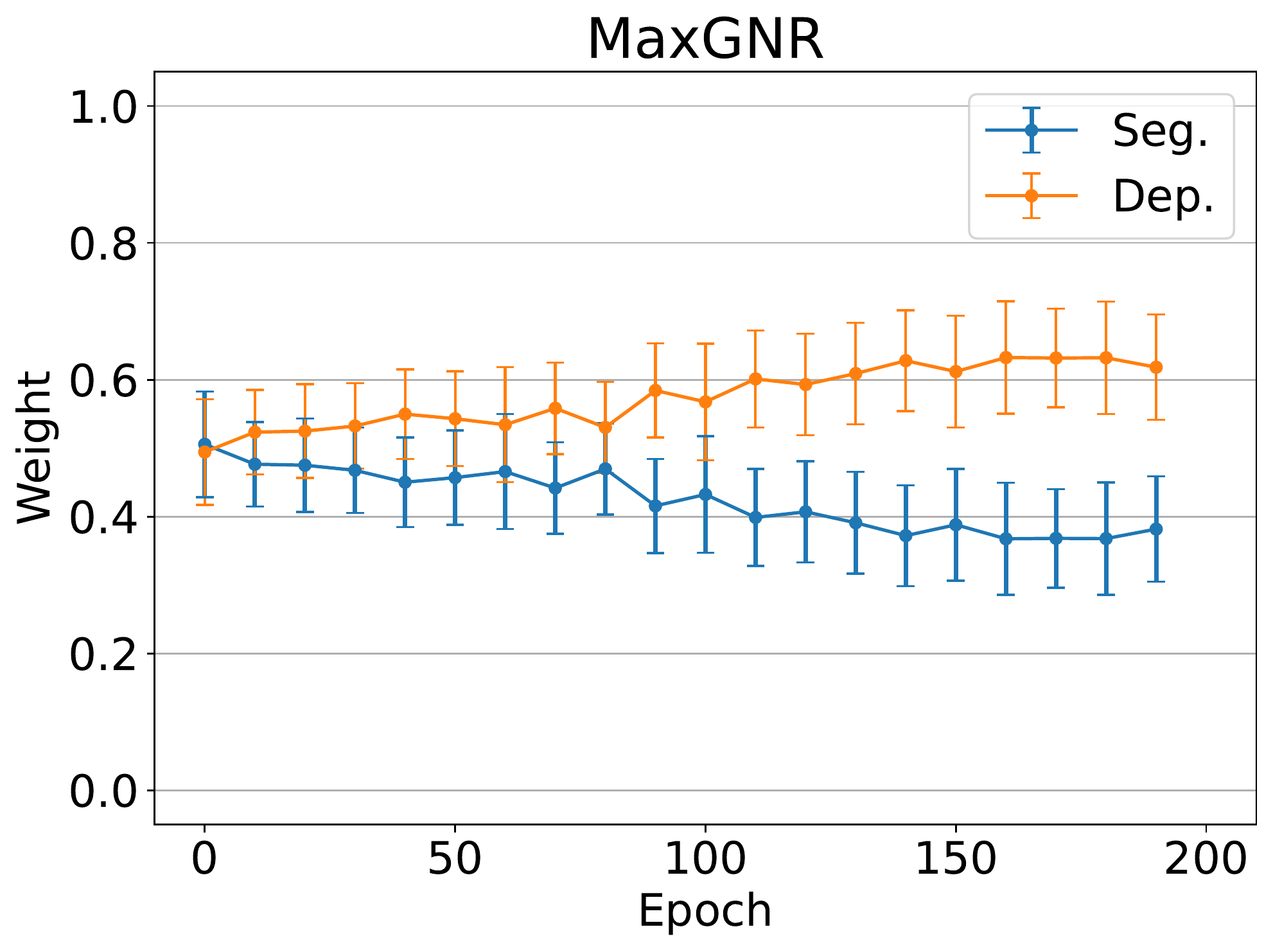}
    \includegraphics[width=0.35\columnwidth]{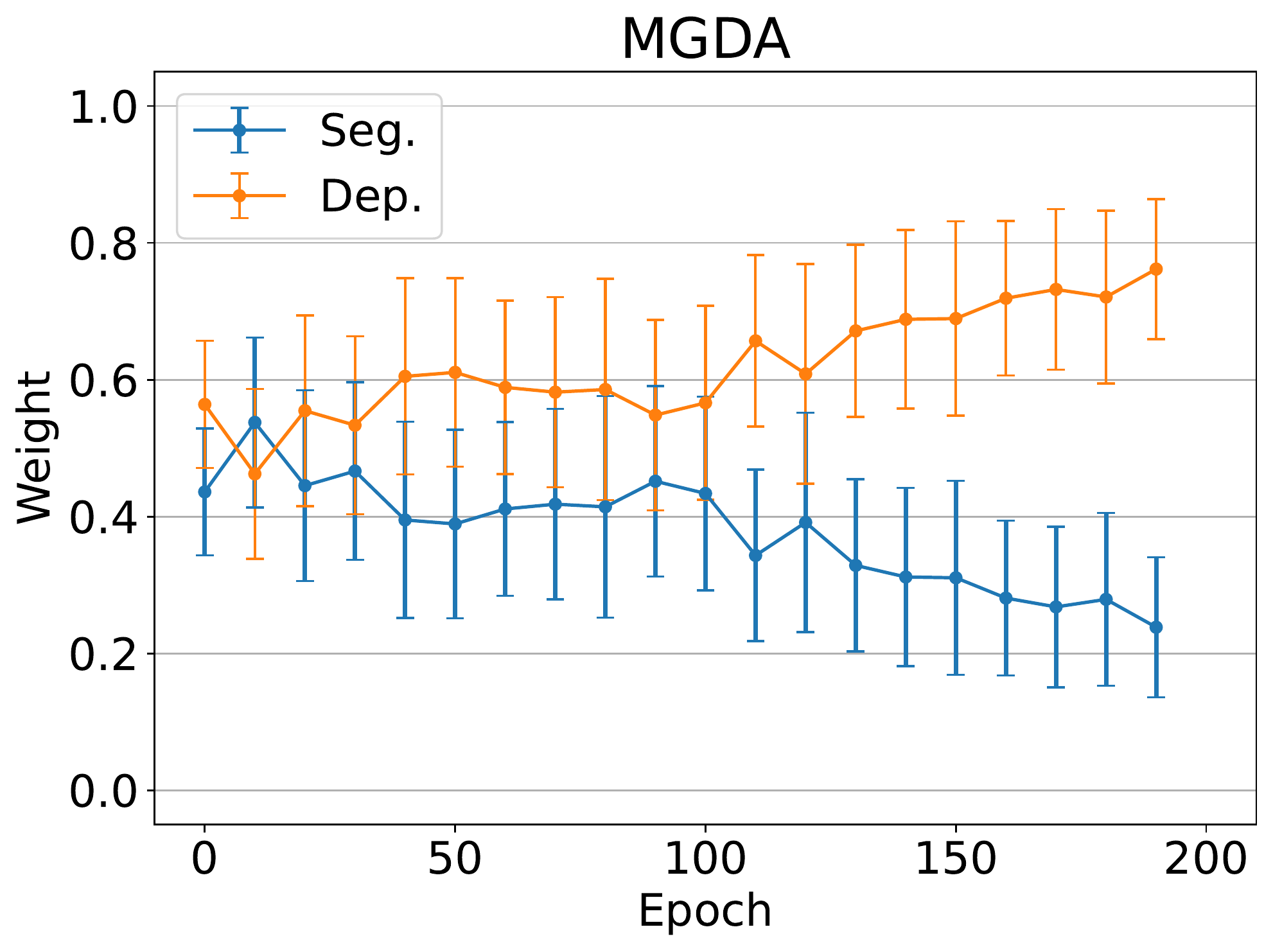}
    \label{f5-2b}
    }
    \caption{Dynamic change of weights and the training process. }
    \label{f5-2}
\end{figure}

The dynamic weight strategy can assign appropriate weights to each task at different training stages. We compared the weights of the different stages obtained by MaxGNR and MGDA in Fig. \ref{f5-2}. We found that the weight trends obtained by the two algorithms were similar, but MGDA's weights were more extreme compared to MaxGNR's weights. Specifically, in NYUv2, the weight assignment of MGDA led the model to focus only on the Surface Normal task, while MaxGNR's assignment was intuitively more balanced. The experimental results (MGDA in Tab. \ref{tab2}) showed that extreme weight assignments may lead to insufficient training problems. The situation was similar in CityScapes. We believed that the weight assignment difference came from the object being balanced: MGDA attempted to balance each task's gradient, while MaxGNR attempted to balance each task's GNR. 



\section{Discussion}

\subsection{Gradient Noise and Performance}

\begin{figure}[t]
    \centering
    \subfigure[NYUv2]{
    \includegraphics[width=0.67\columnwidth]{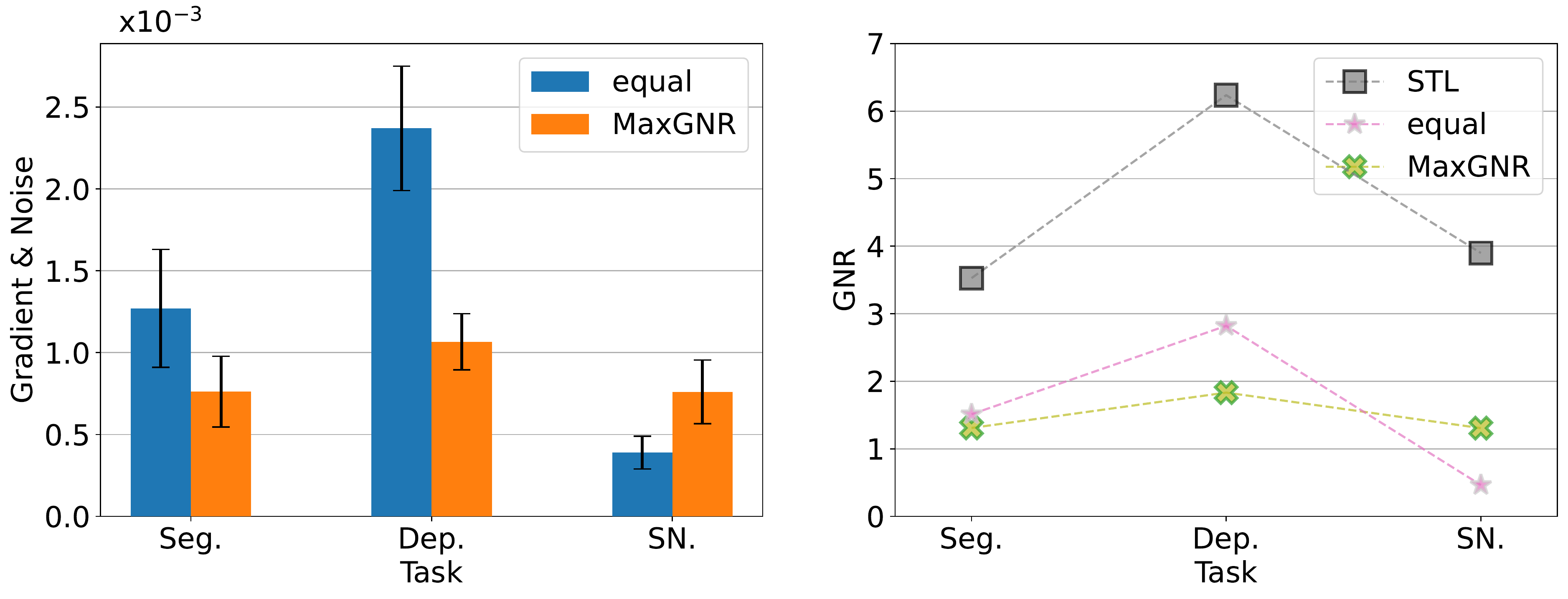}
    \label{f5-1a}
    }
    \subfigure[CityScapes]{
    \includegraphics[width=0.67\columnwidth]{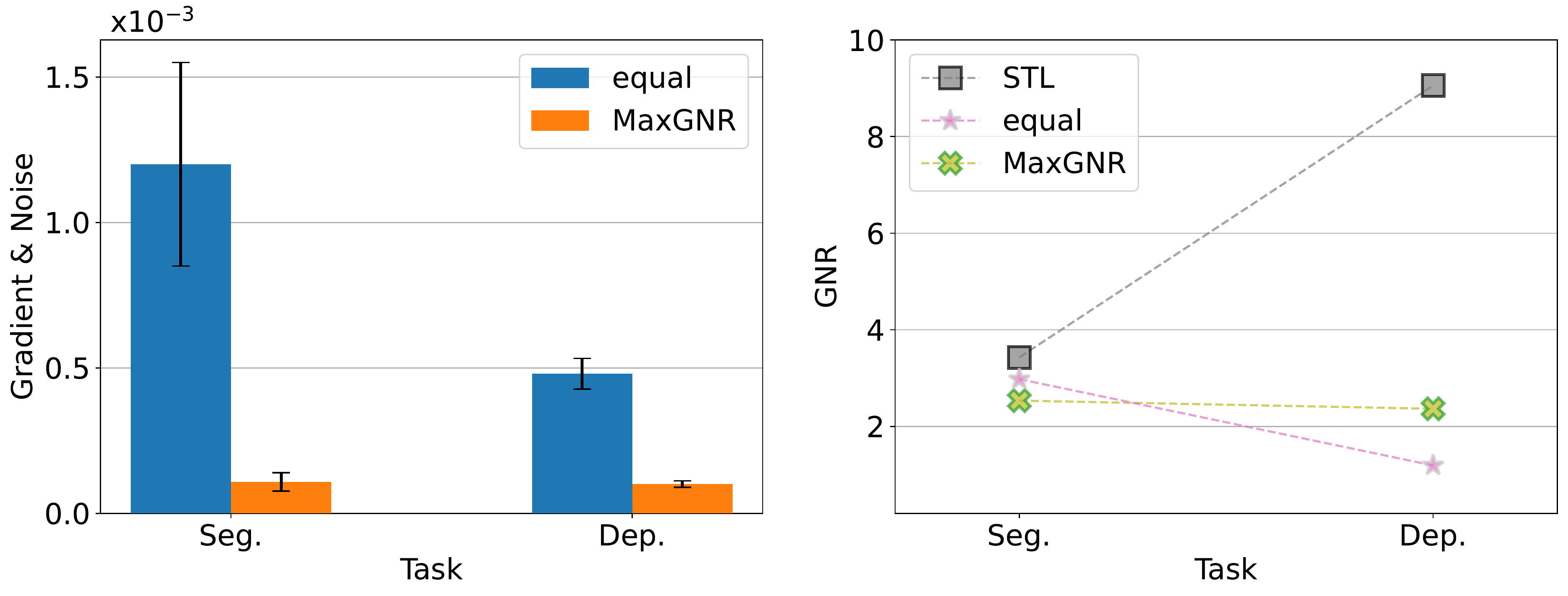}
    \label{f5-1b}
    }
    \caption{Gradient Norm Distribution and GNR of each task in NYUv2 and CityScapes. }
    \label{f5-1}
\end{figure}

We fixed the best models under equal weights and MaxGNR settings and counted the distribution of the gradient norm for all data and calculated the current GNR of the models based on the gradient norm distribution, as shown in Fig. \ref{f5-1}. 

We found that in the equal weights model, the gradients and noises of each task differed significantly, and our algorithm clearly balanced the gradients and noises of each task (Fig. \ref{f5-1} Left). Compared with $\text{GNR}_S$, the $\text{GNR}_M$ of Surface Normal task in NYUv2 and Depth task in CityScapes decreased most significantly in the equal weights model, and the performances of these tasks also exhibited a huge decrease. Our MaxGNR algorithm balanced each $\text{GNR}_M$ (Fig. \ref{f5-1} Right), by slightly reducing the $\text{GNR}_M$ of some tasks (usually harmless), the tasks suffering from performance deterioration would mitigate the interference of ITGN, and gained performance improvements. The experimental results also showed that the model in the MaxGNR setting had significant performance improvements in the Surface Normal task in NYUv2 and the Depth task in CityScapes. At the same time, the performance improvement of some tasks can help the model to get better representation, which can improve the performance of the model in general. 

\subsection{The Paradox of Weight Design}

In the field of weight design of MTL, researchers frequently constructed optimization algorithms depending on the difficulties of the tasks, however, the starting points of the design sometimes conflicted with each other. For example, Dynamic Task Prioritization (DTP) \cite{guo2018dynamic} allocated ``difficult'' tasks higher task-specific weights, while Uncertainty \cite{cipolla2018multi} assigned ``easy'' tasks higher task-speciﬁc weights. \cite{vandenhende2021multi} gave a qualitative explanation for this conflict paradox: Uncertainty seemed to be more suitable for noisy labeled data, while DTP was more suitable for clean ground-truth annotations. The paradox can be described more clearly when we viewed this phenomenon from the perspective of \text{MaxGNR} algorithm. According to the Eq. \ref{e4-5}, there are two factors that influence task's difficulty: small expected gradient $g(\theta)$ and large variance $\frac{C}{\left| S \right|}$. 

If the model's upper limit is restricted by the small expected gradients of some tasks, the weights of these tasks should be adjusted to expand the expected gradients, similar to DTP. The noise part of Eq. \ref{e4-5} can be ignored as: 
\begin{equation}
    \{\omega_k\} = \argmax_{\{\omega_k\}} \left\{ \min \left\{ \Vert \omega_k g^k(\theta)\Vert^2 \right\} \right\}
   \label{e5-1}
\end{equation}

On the contrary, if the model's upper limit is restricted by the large variances of specific tasks, the weights of these tasks should be reduced, similar to Uncertainty, to reduce the impact on other small gradient tasks. So the gradient part of Eq. \ref{e4-5} can be ignored as: 
\begin{equation}
    \{\omega_k\} = \argmax_{\{\omega_k\}} \left\{ \frac{1}{\sum_{k=1}^n \omega_k^2 tr (C^k) /\left| S \right|} \right\}
   \label{e5-2}
\end{equation}

In conclusion, the \text{MaxGNR} algorithm not only considers the magnitude of the gradient but also considers the noise interference, which to some extent unifies these two kinds of algorithms with distinct starting points. 


\section{Conclusion}

In this work, we attributed the insufficient training problem in MTL to the ITGN interference, and we proposed \text{MaxGNR} algorithm, a novel dynamic weight strategy to alleviate this interference. Experiments verified the effectiveness of our algorithm. Looking ahead, gradient noise in MTL is a new field, and we hope to explore the influence of gradient noise on more complex tasks. Besides, how to choose appropriate tasks for joint learning is an open question, and the GNR framework may be a possible research direction. 

\subsubsection{Acknowledgements} This work was supported by the Shanghai Municipal Science and Technology Major Project (2021SHZDZX0102), and the Shanghai Science and Technology Innovation Action Plan (20511102600). 

%
%
%
\bibliographystyle{splncs04}
\bibliography{reference}
%




\end{document}